\documentclass[conference]{IEEEtran}
\IEEEoverridecommandlockouts
\usepackage{cite}
\usepackage{amsmath,amssymb,amsfonts}
\usepackage{algorithmic}
\usepackage{graphicx}
\usepackage{textcomp}
\usepackage{xcolor}
\def\BibTeX{{\rm B\kern-.05em{\sc i\kern-.025em b}\kern-.08em
    T\kern-.1667em\lower.7ex\hbox{E}\kern-.125emX}}
\begin{document}

\title{Temporal Logic Guidance for Action-Only Diffusion Policies with World Models\\
\thanks{This work was supported by a scholarship of the German Academic Exchange Service (DAAD).}
}

\author{
\IEEEauthorblockN{ Moritz Zoellner, Anastasios Manganaris, and Rohan Paleja}
\IEEEauthorblockA{
\textit{Department of Computer Science, Purdue University} West Lafayette, IN, USA \\
\texttt{\{zoellner, amangana, rpaleja\}@purdue.edu \vspace{-0.27cm}}
}
}

\maketitle

\begin{abstract}
Diffusion policies enable multimodal robot behavior but offer limited ability to choose among behavior modes at inference time, even though such control is desirable in human-robot settings. Prior solutions to this lack of control have utilized Signal Temporal Logic (STL) to express human intentions and provide corresponding guidance for diffusion policy inference. However, these approaches can only guide diffusion policies that jointly generate future actions and states, increasing both complexity and runtime. We propose a novel guidance method for action-only diffusion policies that uses a separate learned world model to enable differentiable evaluation of STL robustness, with its gradient then injected into the diffusion process. This steers behavior toward constraint satisfaction without retraining, improving constraint adherence while preserving task performance. On the \emph{Can Transport} task from Robomimic, our method maintains 100\% task success while reducing constraint violations from over 80\% for baseline methods to 4\%. We also discuss extensions toward improved robustness and more complex constraints.
\end{abstract}


\section{Introduction}

For robots to operate effectively in human environments, including those involving non-expert users, they must be able to adapt their behavior to individual preferences \cite{natarajan2023human, paleja2020interpretable}. As robots become increasingly integrated into personal spaces, such preferences can vary significantly across users and situations. Consider a household robot loading a dishwasher: a cautious user might prioritize the safety of delicate glassware, instructing the robot to \textit{``space the dishes far apart so they don't break,"} while another may value efficiency, demanding it \textit{``stack them tightly to fit as much as possible."}

Since the same robot policy is deployed across different users, it must accommodate various preferences without relying on costly retraining. Modern robot learning methods, such as diffusion policies, learn from diverse demonstrations and can represent multiple valid ways of performing a task, taking in examples from a distribution of intended behaviors and outputting actions that imitate those examples ~\cite{ho2020denoising, chi2025diffusion}. As a result, desired behaviors may already exist within the learned data distribution, but a fixed policy will not be able to reliably select them. This motivates inference-time guidance mechanisms that steer policy behavior without retraining.

Methods for inference-time guidance of diffusion models typically rely on a user's preference expressed as a guidance objective against which sampled action trajectories are optimized. Prior work has shown guidance to be successful using guidance objectives based on classifier models \cite{dhariwal2021diffusion}, arbitrary auxiliary networks \cite{bansal2023universal}, visual observations from goal-states \cite{du2025dynaguide}, and STL expressions \cite{zhong2023guided, meng2024diffusionpolicystl, feng2024ltldog}. STL \cite{maler2004stl} offers a structured and expressive way to specify complex tasks and safety constraints~\cite{manganaris2026formal, kapoor2025stlcg++}. Furthermore, it is comparatively well suited for human-robot interaction as it supports a natural language interface for users, with prior work demonstrating the feasibility of mapping language to STL specifications~\cite{liu2022lang2ltl, liu2024lang2ltl, he2022deepstl, mohammadinejad2024systematic, hurley2024stl}. However, existing methods for STL-based diffusion guidance rely on predicted states during diffusion to evaluate constraints. Producing these states requires joint action-state generation, increasing complexity and training cost \cite{janner2022planning}. Many modern policies generate only action sequences, making such approaches not directly applicable.

\begin{figure}
    \centering
    \includegraphics[width=1\linewidth]{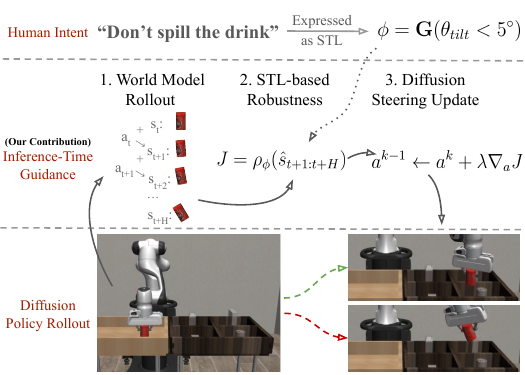}
    \caption{\textbf{Our method} guides the behavior of diffusion policies at inference-time to satisfy STL expressions. We use a world model to roll out the action chunk created by the policy and compute an STL robustness value based on the predicted future states. With all steps being differentiable, we can incorporate the gradient of this robustness value in the denoising process. Our approach allows users to define arbitrary STL constraints at runtime, enabling interactive steering of robot behavior without retraining.}
    \label{fig:method-overview}
    \vspace{-0.9cm}
\end{figure}

We propose an approach, illustrated in Figure~\ref{fig:method-overview}, for inference-time STL guidance of diffusion policies that only produce actions. We use a separately trained, differentiable world model to predict future states of proposed actions, enabling constraint evaluation without requiring state predictions from the policy. This allows our method to work with the majority of existing diffusion policies. It also allows the horizon over which STL satisfaction is evaluated to differ from the horizon over which the diffusion policy predicts actions.
\section{Method}

We consider a diffusion policy $\pi(a_{t:t+H} \mid s_t)$ that generates an action sequence of horizon $H$ conditioned on the current state $s_t$. Let $\phi$ denote a STL specification over state trajectories. Our objective is to guide the sampled action sequence such that the induced trajectory satisfies $\phi$, without retraining the policy. Our approach builds on inference-time guidance of the diffusion denoising process~\cite{ho2020denoising, bansal2023universal}. Starting from noisy actions $\mathbf{a}^k := a_{t:t+H}^{k}$ at denoising step $k$, the policy predicts a cleaner action sequence via
\begin{equation}
\mathbf{a}^{k-1} = \mu_\theta(\mathbf{a}^{k}, s_t, k) + \sigma^k \epsilon.
\label{eq:reverse_step}
\end{equation}
Here, $\mu_\theta$ denotes the denoising model, $\sigma^k$ is the variance for the $k$-th denoising step, and $\epsilon \sim \mathcal{N}(0, I)$. Inference-time guidance modifies this update by adding the gradient of a guidance objective $J$ with respect to the action sequence
\begin{equation}
\mathbf{a}^{k-1} = \mu_\theta(\mathbf{a}^{k}, s_t, k) + \lambda \nabla_{\mathbf{a}^k} J(\phi, \mathbf{a}^{k}, s_t) + \sigma^k \epsilon .
\label{eq:diffusion_guidance}
\end{equation}
Here, $\lambda$ controls the guidance strength. We additionally apply a small number of gradient ascent steps on $J$ to the final action sequence $\mathbf{a}^0$, which we find improves constraint satisfaction.

The guidance objective $J(\phi, a_{t:t+H}, s_t)$ is based on the robustness measure for an STL specification $\phi$, the proposed action sequence $a_{t:t+H}$, and the current state $s_t$. The robustness measure, denoted $\rho_\phi(\tau)$, quantifies the degree to which a trajectory satisfies $\phi$. The resulting objective is given by
\begin{equation}
J(\phi, a_{t:t+H}, s_t) = \rho_\phi(\hat{s}_{t+1:t+H}).
\label{eq:guidance_objective}
\end{equation}
Here, $\hat{s}_{t+1:t+H}$ denotes a prediction of the future state trajectory induced by the action sequence. To obtain this trajectory, we employ a separately trained world model $\hat{s}_{t+1} = f_\theta(s_t, a_t)$, learned from the same data used by the diffusion policy, that approximates the next state given the current state and action. By iteratively applying this model over the action sequence, we obtain $\hat{s}_{t+1:t+H}$. Since the STL robustness is differentiable with respect to a state trajectory, and the state trajectory is differentiable with respect to the input actions through the world model, we can compute the gradient of $J$ with respect to the action sequence and inject it into the diffusion denoising process, thereby steering the policy toward constraint-compliant behavior at inference time.

\section{Results}

\begin{figure}
    \centering
    \includegraphics[width=3.5in]{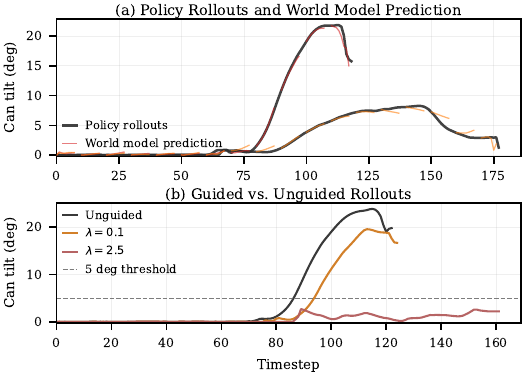}
    \caption{\textbf{Top:} The diffusion policy exhibits multiple behaviors for the same initial state. The learned world model accurately predicts multi-step rollouts over the action horizon. \textbf{Bottom:} When the policy selects an undesirable mode, our method can guide it toward trajectories that satisfy the constraint.}
    \label{fig:results}

    \vspace{-0.2cm}
\end{figure}

We evaluate our approach on the \textit{Can Transport} task from Robomimic~\cite{robomimic}, where a robot must grasp a can and drop it into a box, using a diffusion policy trained on mixed human demonstrations to capture diverse behaviors. We constrain the policy to keep the can upright with the STL specification $\mathbf{G}(R_{zz} > \cos(5^\circ))$, where $R_{zz}$ denotes the alignment between the can orientation and the world $z$-axis. Figure~\ref{fig:results}(a) illustrates that the learned policy exhibits multiple behaviors for the same initial state, including trajectories that violate the uprightness constraint, while the learned world model accurately predicts these rollouts over the action horizon. Figure~\ref{fig:results}(b) shows that our guidance method can steer undesirable policy rollouts toward trajectories that satisfy the constraint.

We evaluate the constraint satisfaction rate over 50 rollouts and compare against the base diffusion policy and a sample-and-rank baseline~\cite{qi2025strengthening} using the same robustness objective. As shown in Table~\ref{tab:results}, our method improves constraint satisfaction while maintaining task success, outperforming both baselines. This suggests that gradient-based guidance more effectively steers the policy toward constraint-compliant behaviors than sampling-based selection, potentially recovering underrepresented modes in the learned distribution.

\begin{table}[t]
\centering
\caption{Constraint satisfaction and task success.}
\begin{tabular}{lccc}
\hline
\textbf{Method} & \textbf{Avg. Tilt ($^\circ$) $\downarrow$} & \textbf{Succ. (\%) $\uparrow$} & \textbf{Viol. (\%) $\downarrow$} \\
\hline
Base Policy & 8.51 & 100.0 & 84.0 \\
Sample \& Rank & 8.42 & 98.0 & 82.0 \\
Guidance (Ours) & \textbf{1.93} & \textbf{100.0} & \textbf{4.0} \\
\hline

\end{tabular}
\label{tab:results}
\vspace{-0.35cm}
\end{table}

\section{Future Work}

We plan to evaluate our approach across more tasks, environments, and STL specifications to better understand its generality. A promising direction is improving stability or efficiency of the guidance process using evolutionary or second-order optimization methods. The most significant direction for our approach is toward supporting \emph{long-horizon} STL constraints, which is a limitation of most existing STL-guidance approaches \cite{zhong2023guided, meng2024diffusionpolicystl, feng2024ltldog, kapoor2026safedec}. A separate world model allows evaluating the consequences of chosen actions at the abstraction level provided by the STL specification's automaton representation \cite{baier2008principles}. Instead of predicting the next environment state given the current environment state and proposed action, we can learn to predict the next automaton state given the current environment state, automaton state, and action mode chosen by the policy. We view our current approach as one step toward this idea to enable significantly longer term, specification-relevant prediction, which will be necessary for finally supporting diffusion policy guidance with arbitrarily complex task specifications.

\clearpage

\bibliographystyle{IEEEtran}
\bibliography{references}

\end{document}